\documentclass[letterpaper, 10 pt, conference]{ieeeconf}  

\IEEEoverridecommandlockouts                              

\usepackage{graphics} 
\usepackage{epsfig} 

\title{\LARGE \bf
A model for traffic incident prediction using emergency braking data
}

\author{Alexander Reichenbach$^{1}$ and J.-Emeterio Navarro-B.$^{2}$
\thanks{$^{1}$Alexander Reichenbach was with MBition GmbH, Daimler AG and is now a graduate student at the Department of Computing,
        Imperial College London, South Kensington Campus, London SW7 2AZ, UK
        {\tt\small alexander.reichenbach20@imperial.ac.uk}}%
\thanks{$^{2}$Jesús-Emeterio Navarro-Barrientos is with MBition GmbH, Daimler AG,
        Dovestrasse 1, 10587 Berlin, Germany
        {\tt\small jesus\_emeterio.navarro-barrientos@daimler.com}}%
}

\usepackage{fancyhdr}
\usepackage{lipsum}

\fancypagestyle{specialfooter}{%
  \fancyhf{}
  
  \fancyfoot[C]{
\small © 2021 IEEE.  Personal use of this material is permitted.  Permission from IEEE must be obtained for all other uses, in any current or future media, including reprinting/republishing this material for advertising or promotional purposes, creating new collective works, for resale or redistribution to servers or lists, or reuse of any copyrighted component of this work in other works.
}
}

\begin{document}

\maketitle
\thispagestyle{specialfooter}
\pagestyle{empty}

\begin{abstract}

This article presents a model for traffic incident prediction.
Specifically, we address the fundamental problem of data scarcity in road traffic accident prediction by training our model on emergency braking events instead of accidents.
Based on relevant risk factors for traffic accidents and corresponding data categories, we evaluate different options for preprocessing sparse data and different Machine Learning models. 
Furthermore, we present a prototype implementing a traffic incident prediction model for Germany based on emergency braking data from Mercedes-Benz vehicles as well as weather, traffic and road data, respectively.
After model evaluation and optimisation, we found that a Random Forest model trained on artificially balanced (under-sampled) data provided the highest classification accuracy of 85\% on the original imbalanced data.
Finally, we present our conclusions and discuss further work; from gathering more data over a longer period of time to build stronger classification systems, to addition of internal factors such as the driver's visual and cognitive attention.

\end{abstract}

\section{INTRODUCTION}
On January 29, 1886, the German engineer and inventor Carl Benz applied for a patent for a “vehicle powered by a gas engine”.
This vehicle, later called the “Benz Patent Motor Car”, was the first ever automobile.
The continuous industrialisation and technological evolution of the world improved the capabilities of motorised vehicles to travel to speeds never thought before.
At the same time, this development has given a new level of relevance to an issue, which has been around long before Carl Benz received his famous patent and is still omnipresent today: {\it road traffic accidents}.
Currently, every year, about 1.3 million people are killed and more than 50 million severely injured in traffic accidents worldwide \cite{ITF2017}.
To counteract this situation, governments, researchers and engineers around the world are trying to reduce the number of fatalities on the road.
In Germany, for example, throughout the past 25 years, the number of traffic accident fatalities has been reduced by 70\%.
However, the number of injury crashes between 1991 and 2015 decreased only by 20,6\%.
For example, in 2015, there were still roughly 3,500 fatal and over 300,000 injury traffic accidents in Germany~\cite{ITF2017}.

These statistics show that there is still potential to improve road safety.
The goal of this article is to propose a model for traffic incident prediction.
To achieve this, we used methods of Data Analytics and Data Mining to build a system capable of: i) interpreting a snapshot of quantifiable circumstances of a vehicle on the road, and ii) producing a prediction of the likelihood of an impending traffic incident.
In the following, we present the current state of the art in the field of traffic incident prediction.
Afterwards, we describe the main concept and the prototype we built for our use case, Germany, based on the previous work by one of the authors~\cite{Reichenbach2018}.
Finally, we present our conclusions about the developed concept and its practical usability, as well as the further work.

\section{State of the art}

The digitalisation of the world and the implicit growing availability of data and computational power have led to the rise of Data Mining for road traffic accident prediction models (APMs).
In 2010, the Highway Safety Manual (HSM) was published by AASHTO \cite{AASHTO2014}, as a comprehensive high-level guide to traffic accident prediction models.
In recent years, most of the developed models that can be found in academic literature focus on specific regions, such as Finland \cite{Turunen2017} or Iowa, USA \cite{Yuan2017}.
Some other studies focus on certain types of accidents only, like accident severity in Seoul \cite{Lee2018}, motorway accidents in England \cite{Michalaki2016} or car-cyclist accidents in different US states \cite{Asgarzadeh2018)}.
While researchers try to obtain a generalised transnational APM, it currently seems unlikely that one single APM with one defined set of features~/~explanatory variables is capable of universally provide valid traffic accident prediction \cite{LaTorre2016}.

\subsection{Relevant Input Data for Road Traffic Accident Prediction Models}
\label{sec:RelevantInput}

To analyse and predict road traffic accidents, it is necessary to view the main factors contributing to road crashes.
In 2004, the World Health Organization (WHO) has conducted a comprehensive study on Road Traffic Injury Prevention \cite{Peden2004}, which identified the following key risk factors influencing road accident involvement:

{\it Speed} is identified as one of the most important risk factors for road traffic accident involvement as well as accident severity in 2004 and 2015.
The probability of a crash involving an injury is proportional to the speed squared \cite{Peden2004}.
{\it Driver age and experience} is another significant factor, especially young drivers \cite{Peden2004}, and senior citizens above 65 years of age \cite{ITF2017} pose a growing threat.
{\it Driver impairment} through intoxication or fatigue represents a major risk. Intoxication remains one of the main concerns \cite{Peden2004,WHO2015}.
{\it Inadequate visibility} plays a significant role for accident involvement, which can be caused by missing road lighting or poor weather conditions.
{\it Road- and vehicle-related factors} can be key factors for traffic accidents as well. While the previously mentioned factors are all (except for inadequate visibility) exclusively influenced by human factors, the external conditions of the roads and vehicles also largely contribute to accident involvement under certain circumstances.
Some road sections are especially dangerous due to their construction and certain vehicles that are equipped with special safety features are capable of drastically reducing the risk for traffic accident involvement \cite{Peden2004}.

A model to predict road traffic accidents should therefore be given as inputs any available data related to these aforementioned factors.
However, data availability is the first big challenge.
While for example road, traffic and accident data is generally highly available in European countries, the availability of data about the driver and their current state or behaviour is very limited \cite{Yannis2016}.

The fact that road traffic accidents are rare events leads to the problem of {\it class imbalance}, which has the potential to bias a prediction model.
To address this issue, an analysis algorithm can be chosen, which is specifically incentivised to produce accurate predictions for the minority class (i.e. traffic accidents), such as \cite{Theofilatos2016}.
Alternatively, it is also possible to address the problem by modifying the data underlying the model through e.g. over- or under-sampling (to equalise sample sizes for each class) or bootstrapping (to artificially increase the overall amount of available data). 
Another issue is the {\it spatial variance}, there is significant difference between accidents in urban and rural areas, resp., this because of the overall traffic behaviour changes depending on the surrounding road situation~\cite{LaTorre2016}.

\subsection{Algorithms for traffic accident prediction}

The problem at hand - predicting whether under given circumstances a road traffic accident is likely to occur - is a classical binary classification problem.
While classical approaches such as the Highway Service Manual \cite{AASHTO2014} focus largely on linear models for traffic accident prediction, recent research suggests stronger classification accuracy for models based on non-linear correlations.
As such, Machine Learning methods like Artificial Neural Networks, Decision Trees, Random Forests and Support Vector Machines have been used by researchers to predict road traffic accidents \cite{Yuan2017,Ren2018,Wenqi2017}.
Classical statistical methods, especially Logistic Regression or Probit models can  also be found in recent research \cite{Mannering2014}.
For this paper, we have selected the following algorithms for evaluation: Logistic Regression as a “classical” approach to traffic accident prediction, tree-based algorithms (Decision Trees and Random Forests) and Neural Networks.

\section{CONCEPT}
This section proposes a general approach for the implementation of a system capable of predicting Road Traffic Incidents using methods of Machine Learning.
In this work we have chosen to follow the CRISP Data Mining framework \cite{Shearer2000} to build a system that can eventually be taken to production. The requirements for our system are to first receive a request sent by a specific vehicle which contains the current position of the vehicle.
The prediction system then retrieves the current external conditions of the vehicle at its given position and passes these along to its prediction model as input variables.
The model then produces a prediction of the likelihood of a traffic indent under the current circumstances of the vehicle.
Finally, the result is sent back to the vehicle as a response to its request.

\subsection{Data understanding}

The dependent variable is already defined to be the binary value corresponding to either an instance of an accident or not an accident.
One important assumption in this paper is that we consider emergency braking data as near-accident data.
This significantly increases the amount of available data.
As emergency braking however cannot entirely be treated as equal to real accidents, the system we propose is referred to as a {\it Traffic Incident Prediction} system, as opposed to a {\it Traffic Accident Prediction} system, which is based on confirmed accident data.
To our knowledge, no research paper has approached Traffic Incident Prediction this manner.

Modern automobiles are equipped with advanced emergency braking systems capable of the identification of emergency braking versus regular use of the brake.
Concretely, we use for our analysis emergency braking data from {\it Car2X-Communication} systems, see CAR-2-CAR Communication Consortium Manifesto ({\it https://www.car-2-car.org}) for more information, which provides emergency braking events available to other vehicles near-by.
On top of the actual emergency braking data, negative instances, i.e. non-emergency braking events, are also required in order to adequately construct a prediction model.
This concept is therefore based on vehicle status data in the form of events, which consist of binary information either representing an emergency braking or any other situation of a vehicle on the road at a specific location and point in time.

\begin{figure}[thpb]
   \centering
   \framebox{\parbox{3.3in}{
   \includegraphics[width=3.3in]{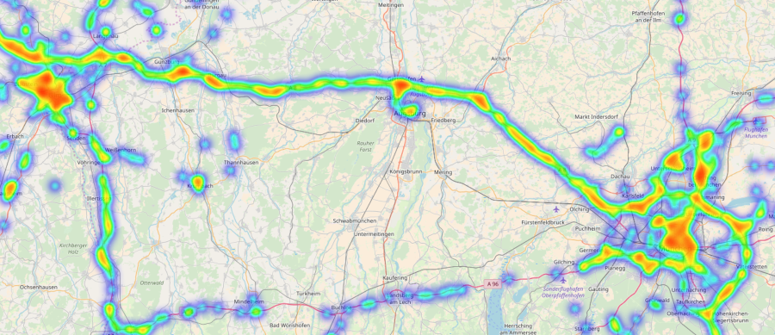}
   }}
   \caption{Heatmap of emergency braking events between Stuttgart and Munich shows aggregations on major roads.}
   \label{fig:Heatmap}
\end{figure}

We analysed the Car2X event data sent by vehicles (emergency braking or regular events from May to July 2018) using a heatmap plot, see Fig.\ref{fig:Heatmap}.
Certain idiosyncrasies can be seen from its spatial distribution.
Large amounts of data originated at the exact same positions (as for example the Sindelfingen production facility), which are not suitable for inclusion into the dataset which is finally to be analysed.
For this reason, the map visualisation was examined in more detail and certain areas of unrealistically high event aggregation were recorded.

We choose emergency braking data as the dependent variable for our traffic incident prediction model.
The source data for this is acquired from {\it Mercedes-Benz vehicle’s Car2X-Communication data}.
This data contains messages which are sent to Daimler Servers in case a vehicle has detected an emergency braking.
Emergency brakings are defined as situations in which the vehicle detects high deceleration over a continuous amount of time at a speed higher than the regular inner-city speed limit.
These constraints ensure that every emergency braking event message can be assumed to be a real emergency braking and not just a situation in which the driver accidentally applied the brake too strongly.
Any data that was used for this prototype was pseudo-anonymised without any means available to reconnect a single driver or vehicle with a sent Car2X message.

Based on the categories of input feature data which were suggested in sec.\ref{sec:RelevantInput}, the following further data sources were selected:
\begin{itemize}

\item {\it Weather data} was acquired through a big weather data provider based in Berlin (more details are not disclosed due to confidentiality reasons).
This data contains information about temperature, precipitation, air pressure and visibility.
\item {\it Road and traffic data} was acquired from another data provider which cannot be named due to legal requirements.
The traffic data contained information on the current, monthly average and so-called reference speed, which refers to the road’s speed limit.
This traffic data is provided in association with street segments, which in turn carry basic information on their affiliated streets like their names, locations, and functional road class.

\end{itemize}

\subsection{Data preparation}

\begin{figure}[thpb]
   \centering
   \framebox{\parbox{3in}{
   \includegraphics[width=3in]{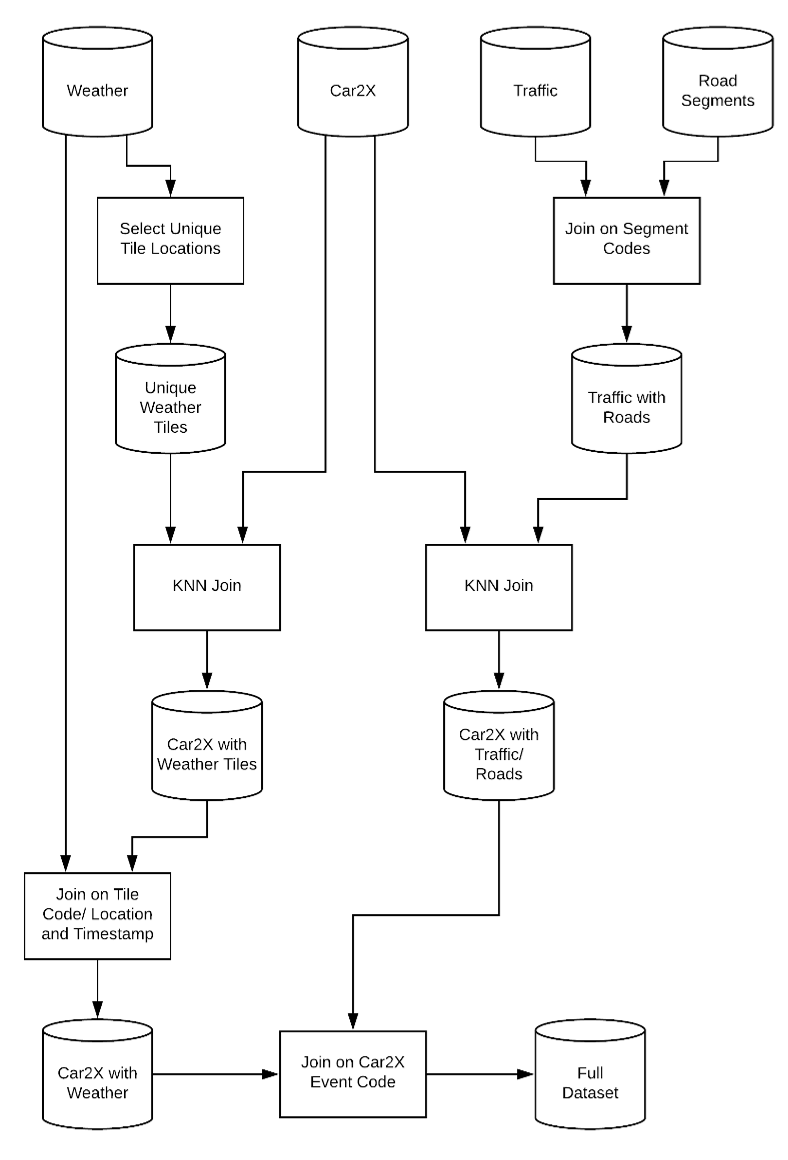}
   }}
   \caption{Data matching process.}
   \label{fig:dataMatching}
\end{figure}

The first step is to gather together all input data (emergency braking, weather, time and date, road and traffic data) in a single dataset.
The major task is to match emergency braking or regular event sent from a vehicle with location and point in time to the feature data.
For this, data are loaded into a geo-database which can then perform the computationally complex spatial matching.
We selected {\it SpatiaLite} ({\it https://www.gaia-gis.it/fossil/libspatialite}) as geo-database for spatial queries.
A {\it Data Integrator} component implemented in Java provides extraction of raw input data (available in JSON format), aggregation/ transformation of the raw data into a structured form and insertion of structured data into the database.
Furthermore, multiple SQL queries are conducted to join the separate datasets into one final database table.
Fig.~\ref{fig:dataMatching} shows an overview of the data matching process.
In this process, the Car2X events have to be matched up with weather, traffic and road data.
Note that traffic data is already affiliated with a road segment via a segment code, these two categories can be matched up with a simple SQL (inner) join.
To match the resulting table (Traffic with Roads) to each Car2X event, the street segment whose center point is the closest to the event position is matched up with the event through a spatial KNN (k-nearest-neighbours) query.

For weather data, an extra step has to be taken to enable automated matching of single batches of weather data to the Car2X table: Before any join operation is executed, all unique weather tiles and their locations are extracted from a sample of the data (which needs to contain all unique tiles).
These unique tiles are then matched up to all Car2X events through another KNN query.
In this manner, the automated matching of the full weather data and the Car2X events can be done on a simple and computationally inexpensive join based on the tile or location.
On a final step, we join the matched weather data with the Car2X with Traffic/Roads table on the Car2X event code.

\section{MODEL IMPLEMENTATION}
For data manipulation, we used the following Python libraries: Pandas, NumPy, Matplotlib, Scikit-learn and Imblearn.
The final dataset contains about 110,000 records, each of which represents a single Car2X event with corresponding temporal, natural lighting, weather, traffic and road data.
Of all records, approx. 10\% correspond to emergency braking events.

As a final step to enable the training and evaluation of the potential prediction model(s), the entire dataset is split into training, validation and test subsets. For the purpose of this prototype, a 70\%:15\%:15\% split was chosen.

\subsection{Feature selection}

\begin{figure}[thpb]
   \centering
   \framebox{\parbox{3in}{
   \includegraphics[width=3in]{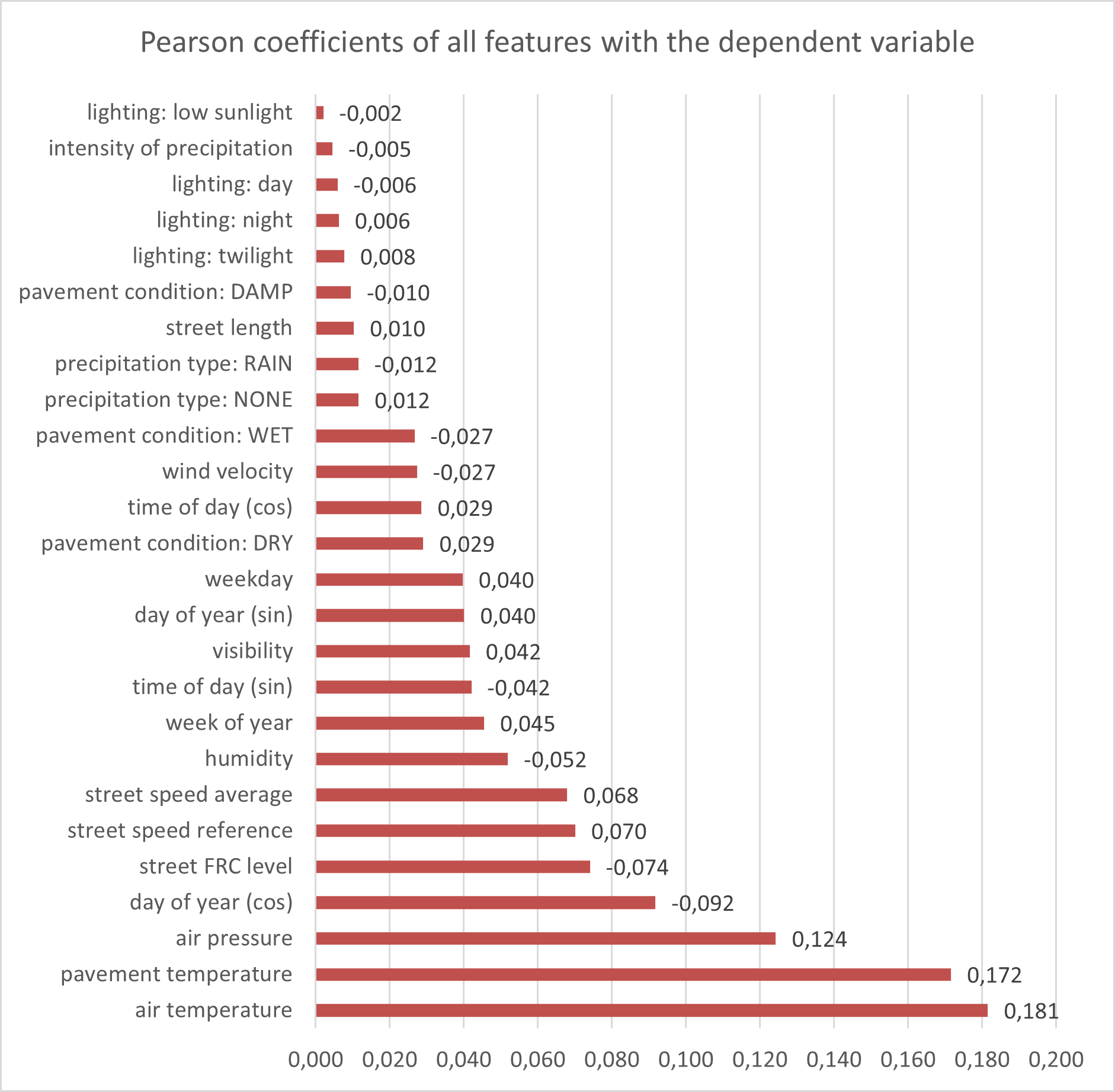}
   }}
   \caption{Correlations of features with the dependent variable, ordered by coefficient absolute values and annotated with the original (signed) values.}
   \label{fig:featuresCorrelationsDependentVar}
\end{figure}

We are interested in correlations between the feature “emergencyBraking" as the dependent variable and all other features.
Fig.\ref{fig:featuresCorrelationsDependentVar} shows an initial overview of all correlations, calculated based on the Pearson Correlation Coefficient, which only reflects correlations of linear nature.
It can be seen that the strongest of these correlations are {\it air temperature ($\rho = 0.181$)} and {\it pavement temperature ($\rho = 0.172$)}
The distribution of the single values of these feature show that emergency braking is more likely to occur at higher temperatures, especially above 20$^{\circ}$C.
At lower temperatures, the relative number of regular events is significantly larger than that of emergency braking events.
{\it Air pressure} is the feature with the third strongest correlation with the dependent variable.
As high air pressure generally indicates “cloud-free and fair weather conditions", this in combination with the air temperature distribution can be interpreted as a predisposition of drivers towards more careless driving behaviour in warm and sunny weather.
While this correlation may seem unintuitive at first, research has found this to be a general trend \cite{Brijs2008}.
We ignored the cosine of the day of the year as it is not directly interpretable and it may not be reliable, given the sample data only represents a fraction of the year.
The {\it street FRC level} is the functional road classification to the respective street (segment).
It is a categorical feature with values: 1) Highways and major intersections, 2) Major artery, 3) Major road, and 4) Neighbourhood street.
Out of the three large FRC levels, comparatively more emergency braking occur on streets of level 1, on level 2 there is roughly an even distribution and on level 3 there are relatively more regular events than emergency braking.
Emergency braking events therefore seem to be most likely on roads of the highest FRC classification (level 1).
However, the lack of emergency braking data on low FRC level roads may be explained by the speed threshold which has to be reached to trigger emergency braking Car2X messages.
Similarly, for {\it street speed reference and average}, we note that on streets with high reference speeds, the relative number of emergency braking events lays above that of regular events, while the opposite is the case for lower reference speeds.
Finally, we note that the feature {\it time of day} shows relatively fewer emergency brakings than regular events during the day (between around 5:00 and 14:00).
For the rest of the day, both about equal, except for the timeframe between 14:00 and 18:00, when the relative number of emergency braking is higher than that of regular events.
Except for {\it day of the year}, all features presented in this exploratory analysis were included into the modelling.

\subsection{Modelling and evaluation}

In addition to the libraries described above, we used the following libraries for modelling and evaluation of the models: $Scipy$, $Tensorflow$ and $Keras$, the latter two especially for Neural Networks.
We trained and evaluated the following algorithms on the original (imbalanced) and artificially balanced (over-sampled, under-sampled and 50\-50-sampled) datasets: {\it Logistic Regression (LR), Decision Tree (DT), Random Forest (RF) and Neural Network (NN)} models.

To measure the performance of the models, we use $ROC$ (Receiver Operator Characteristic) and $AUC$ (Area Under the ROC Curve).
This represents the trade-off between the True Positive Rate (Recall) and the False Positive Rate.
Statistically, the resulting value of the AUC is equal to the probability that a randomly chosen positive instance will be ranked higher (i.e. is more likely to be predicted as positive) than a randomly chosen negative instance \cite{Fawcett2006}.

\begin{figure}[thpb]
   \centering
   \framebox{\parbox{3.3in}{
   \includegraphics[width=3.3in]{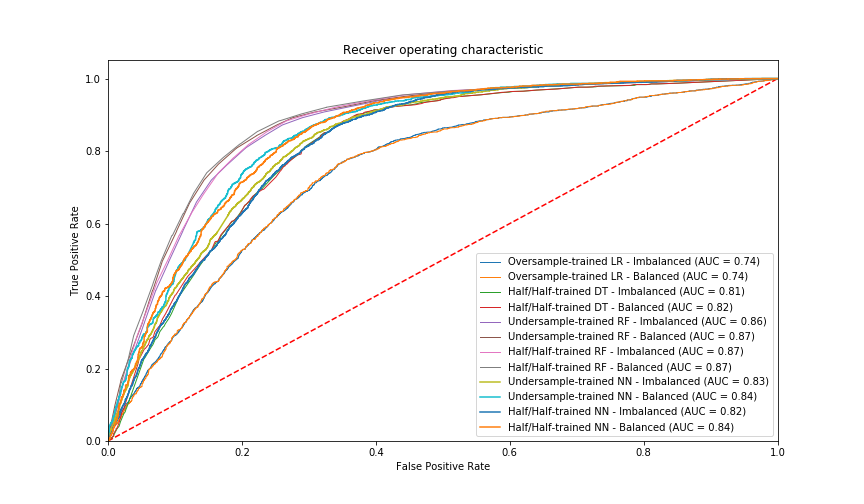}
   }}
   \caption{ROC curves and AUC scores for selected models.}
   \label{fig:ROCnAUC}
\end{figure}

The models with the highest performance were selected for hyper-parameter tuning and re-evaluated.
Finally, the new classification performance scores are compared and the most suitable model is selected.
Depending on the constraints of the concrete implementation, the model’s interpretability as well as the computational complexity of producing new predictions were taken into account for deciding on the best model for our prototype.
Out of all model types, at least one model-sample combination was selected.
Fig.~\ref{fig:ROCnAUC} shows the ROC curves and AUC scores for our selected models (evaluation with respect to the validation dataset).
Note that under-sampling is the best sampling method to produce models which perform well both on the artificially balanced and the original imbalanced data.
Other sampling methods also deliver strong results, while as expected any model trained on the imbalanced dataset did not produce reasonable results.

For the neural network models, we started with the following initial hyperparameters:
\begin{itemize}

\item Network structure: 3 layers with 23, 10 and 1 neurons, resp.
\item Activation functions:	ReLu is used for the first two layers, sigmoid is used for the output layer
\item Loss function:		binary cross-entropy
\item Optimizer:		adam

\end{itemize}

For random forest models, we considered a number of estimators = 25 for our initial hyperparameters.

Fig.~\ref{fig:ROCnAUC} shows a subset of all models trained on the differently sampled datasets.
Generally, the imbalance-trained models (especially for NN and LR) are very close to classifying all instances as regular events: Their accuracy on the balanced test data is only slightly above 50\% and they have low recall values on both test sets.
Out of the models trained on balanced data, the undersample- and the 50-50-trained models have the best performance, while the first has a slightly better recall score and the latter performs more strongly in accuracy and precision.
The oversample-trained models are not far away, but tend to perform poorly on the original imbalanced data which is a sign for overfitting.

Note, that the best AUC score so far is achieved by the half-half-trained Random Forest Classifier (0.87).
Due to the overall strong performance of the RF models, they are selected for further evaluation.
The next best models (in order) are the Neural Network, Decision Tree and Logistic Regression models.
Favoring human-understandable over black-box models and considering the minor difference in performance between the NN and DT models, the DT models are selected for further evaluation as well.

\begin{figure}[thpb]
   \centering
   \framebox{\parbox{3.3in}{
   \includegraphics[width=3.3in]{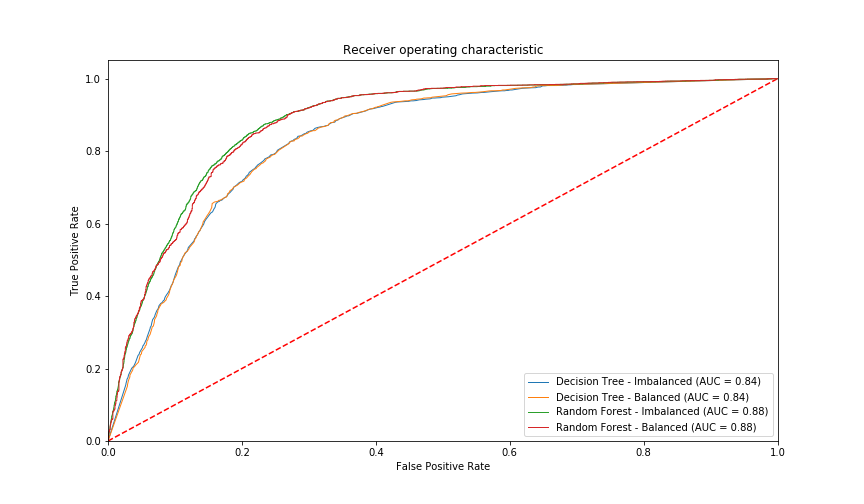}
   }}
   \caption{ROC curves and AUC scores for final models.}
   \label{fig:ROCnAUCfinalModels}
\end{figure}

As a next step, simple hyper-parameter tuning is applied to both selected models using randomized search within a selection of possible hyper-parameter settings.
For each model, 50 combinations of hyper-parameters were generated using scikit-learn’s randomized search (RandomizedSearchCV) and evaluated through 3-Fold-Cross-Validation with accuracy, precision, recall and the AUC as performance scores.
Fig.~\ref{fig:ROCnAUCfinalModels} shows the performance of the two optimised models.
The best model configuration for RF was for example: Maximum number of separate trees = 100, maximum tree depth = 341, minimum number of
samples per decision~=~5, and minimum number of samples per leaf~=~1.

\begin{figure}[thpb]
   \centering
   \framebox{\parbox{3.3in}{
   \includegraphics[width=3.3in]{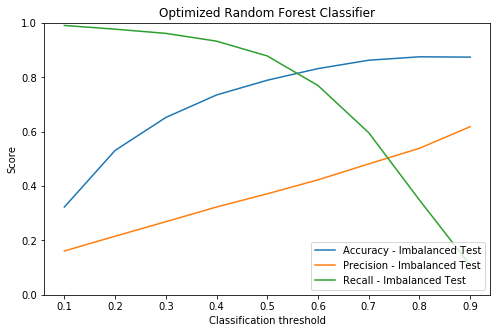}
   }}
   \caption{Performance metrics and classification thresholds for the Random Forest model.}
   \label{fig:OptimzedRFClassifierThreshold}
\end{figure}

As a last step, the development of accuracy, recall and precision in dependence of the classification threshold (the minimum degree of certainty for classification of an instance as emergency braking) is examined in more detail.
Fig.~\ref{fig:OptimzedRFClassifierThreshold} shows the behaviour of three performance metrics for classification thresholds between 0.1 and 0.9 with the default threshold for both models at 0.5.
At this default configuration, for both models accuracy and recall are comparatively high while precision is low.
Approaching the maximum threshold of 0.9, accuracy and precision are increasing while recall is strongly de-creasing.
Depending of the concrete use case of the final traffic incident prediction model, at this point the classification threshold can be adjusted to the point which best fits the business needs of the classification system.
If the main priority for this system is to regardless of their class correctly identify as many events as possible or to ensure the largest possible portion of records that are identified as emergency braking events actually are emergency braking events, the threshold should be increased to gain accuracy and precision.
On the other hand, if the system’s focus is supposed to lay on making sure as many emergency braking events as possible are identified as such, the threshold should be lowered to increase the model’s recall score.

\begin{figure*}[!htp]
   \centering
   \includegraphics[width=14cm,keepaspectratio]{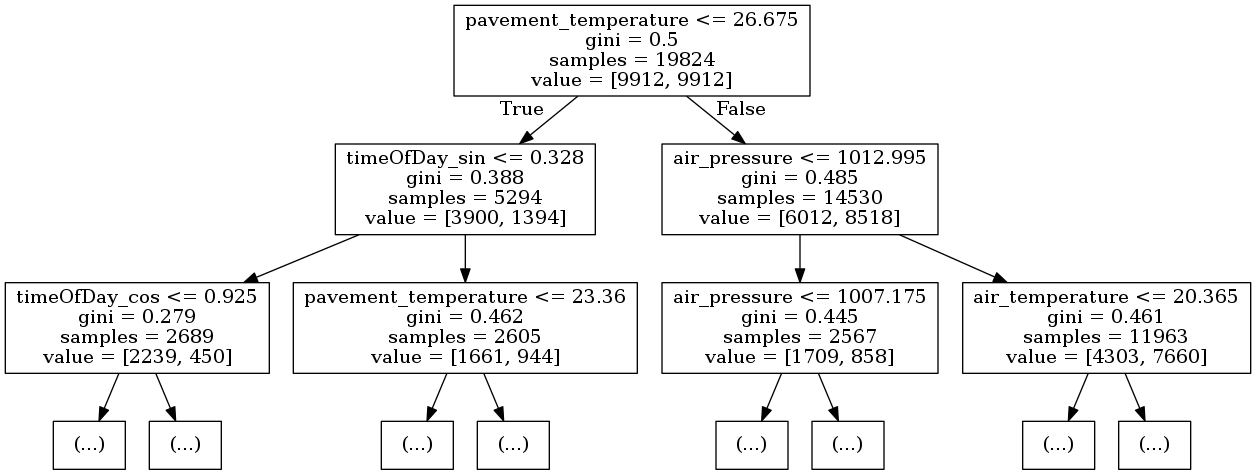}
   \caption{Final decision tree classification model (top layers).}
   \label{fig:finalTree}
\end{figure*}

We chose the Random Forest Classifier with a classification threshold of 0.65 for our prototype, which corresponds to an accuracy score of 84.79\% with recall at 68.50\% and precision at 44.74\%.
To provide better understanding of single classification results, it is additionally proposed that the Decision Tree model (or even a simplified version of it) is utilised as well to provide a “second opinion” for each classification result.
Fig.~\ref{fig:finalTree} shows the first layers of the DT.

\section{CONCLUSIONS AND FURTHER WORK}

Overall, this prototypical implementation of a traffic incident prediction system has produced an explanatory model which is capable of classifying traffic incidents with an accuracy of approx. 85\%.
To achieve this, Car2X, weather, road and traffic data has been collected from different data sources and extensively prepared and explored in order to make it suitable for analysis.
The replacement of accident data with emergency braking data as well as the employment and evaluation of multiple approaches for dealing with the problem of class imbalance have enabled the use of Machine Learning algorithms despite the scarcity of road traffic accident events.
An approach similar to the proposed Traffic Incident Prediction system is not known to the authors at the time of publishing this work.
To further improve on the developed prediction model, future work should evaluate an extension of the categories of input data presented to the learning algorithm. The current model is able to produce fairly accurate predictions considering just the external circumstances of an emergency braking event. Especially driver-centric data such as attention to the road or physical and mental fitness have the potential to further increase model performance.
Further work also includes adjust routing calculation and driving guidance according to the profile of the driver, weather conditions and other external factors. 
For example, if the driver is inexperienced and there is heavy rain, the routing calculating algorithm could avoid intersections, roundabouts, and other road elements where our system predicted the use of emergency braking for the current traffic and environmental situation.


\addtolength{\textheight}{-12cm}   



\section*{ACKNOWLEDGMENT}

Thanks to Frank Bielig and Martin Gebert from MBition GmbH, for their support on gathering and pre-processing the data.
Thanks to Prof. Dr. Andreas Schmietendorf from HWR-Berlin for reviewing first versions of this research work.






\end{document}